# Three-way causal attribute partial order structure analysis

Zaifa Xue[a,b], Huibin Lu[a,b], Tao Zhang[a,b,*], Tao Li[a,b], Xin Lu[a,b]

[a]*School of Information Science and Engineering, Yanshan University, Qinhuangdao 066004, China*

[b]*Hebei Key Laboratory of Information Transmission and Signal Processing, Qinhuangdao 066004, China*

## Abstract

As an emerging concept cognitive learning model, partial order formal structure analysis (POFSA) has been widely used in the field of knowledge processing. In this paper, we propose the method named three-way causal attribute partial order structure (3WCAPOS) to evolve the POFSA from set coverage to causal coverage in order to increase the interpretability and classification performance of the model. First, the concept of causal factor (CF) is proposed to evaluate the causal correlation between attributes and decision attributes in the formal decision context. Then, combining CF with attribute partial order structure, the concept of causal attribute partial order structure is defined and makes set coverage evolve into causal coverage. Finally, combined with the idea of three-way decision, 3WCAPOS is formed, which makes the purity of nodes in the structure clearer and the changes between levels more obviously. In addition, the experiments are carried out from the classification ability and the interpretability of the structure through the six datasets. Through these experiments, it is concluded the accuracy of 3WCAPOS is improved by 1% - 9% compared with classification and regression tree, and more interpretable and the processing of knowledge is more reasonable compared with attribute partial order structure.

**Keywords**: Formal concept analysis, Three-way decision, Attribute partial order structure, Causal inference, Causal factor

## 1. Introduction

Attribute partial order structure analysis (APOSA) is an important method in the field of Concept-cognitive learning (CCL) [4, 31, 32, 19], which explores the relationship between attributes from the perspective of human cognition. Based on formal concept analysis (FCA), attribute partial order structure (APOS) is first proposed by Professor Hong [3], which is an effective cognitive tool

* Corresponding author at: School of Information Science and Engineering, Yanshan University, Qinhuangdao 066004, China.

*E-mail address:* zhtao@ysu.edu.cn (T. Zhang).

for knowledge discovery and knowledge representation. APOSA has been applied in language analysis [29], traditional Chinese medicine [24, 25], etc.

In recent years, Professor Hong has discussed APOS from the perspective of granular computing to realize the knowledge discovery of spleen yang deficiency syndrome in traditional Chinese medicine [13]. Partial order structure and attribute topology are integrated in structural methods [40, 12], and partial order topology has been developed for Parkinson's diagnosis [33]. At the same time, Professor Zhang completes the causal reasoning of Traditional Chinese Medicine syndrome elements based on the partial order expansion of attribute topology [34]. Professor Zhang combines concept tree with attribute topology to form an incremental concept tree to solve the problem of causal asymmetry in computer science [35]. In 2020, Professor Yan integrates partial order structure with three-way decision (3WD) [26] and explores the removal of cross links between object pattern branches in attribute partial order structure diagram (APOSD), and the problem of negative attribute support can be resolved. Although 3WD describes the partial order description from the perspectives of positive domain, negative domain and boundary domain, the structure still belongs to the traditional set coverage research, so his method represents the formal structure of data association.

Studies have shown that causal inference (CI) can eliminate pseudo correlation and explore real causality [6, 2]. CI is first proposed by Professor Pearl in 2009 [15, 16]. Since CI is considered as a necessary tool to realize strong artificial intelligence, more and more scholars have invested in research in recent years. Professor Little proposes three kinds of causal bootstrap algorithms, which can be used for classification and regression [11]. Huawei Noah Ark laboratory applies reinforcement learning to causal discovery algorithm to explore causal structure [41]. In addition, the exploration in the medical field is to find the real cause of diseases and the effect of drug treatment [17, 18, 21], and the exploration in the traffic management field is to predict and reduce the occurrence of traffic accidents [1].

In order to further improve APOS, weaken the correlation between data and mine the essential causality, we combine the method of CI with APOS, and then integrate with the idea of 3WD to explore the construction method of three-way causal attribute partial order structure (3WCAPOS). The main contributions of this work are three-fold:

1. The manuscript introduces the causality analysis into the field of formal concept analysis

and makes set coverage evolve into causal coverage in attribute partial order structure.

2. The causal factor is defined and the related theorems, properties are proved to evaluate the causal coverage between attributes from the view of causality.

3. Three-way causal attribute partial order structure analysis is proposed for supervised learning data based on the causal factor. The feasibility and effectiveness of the proposed method are verified by experiments on six datasets.

The structure of this paper is as follows: Section 1 introduces the research background and purpose. Section 2 briefly introduces some basic concepts of cognitive science and causal reasoning. Section 3 puts forward the CF and the construction method of 3WCAPOS. Section 4 proves the rationality of 3WCAPOS through experiments. Section 5 draws a conclusion.

## 2. Preliminaries

In this section, some basic concepts related to FCA, 3WD, APOS and CI will be briefly introduced.

### 2.1. Formal concept analysis

**Definition 1** In a triple formal context $L = (G, M, I)$, where $G$ is a set of objects, $M$ is a set of attributes, and $I$ is a subset of Cartesian product of $G$ and $M$, and $(g, m) \in I$ indicates the object $g$ has an $m$ attribute, while $(g, m) \notin I$ is the opposite.

Generally, a formal context can be represented by a table with values of 0 and 1. A value of 1 means that the row object has column attributes. Suppose $L = (G, M, I)$ is a formal context, for arbitrary $A \subseteq G$ and $B \subseteq M$.

$$f(A) = \{m \in M \mid \forall g \in A, (g, m) \in I\} \tag{1}$$

$$g(B) = \{g \in G \mid \forall m \in B, (g, m) \in I\} \tag{2}$$

If $f(A) = B$ and $g(B) = A$, then $(A, f(A)) = (g(B), B) = (A, B)$ is called a complete cognitive concept. In a cognitive concept $(A, B)$, $A$ expresses the extension, while $B$ represents the intension. For two formal concepts $(A_i, B_i)$ and $(A_j, B_j)$, where $i \neq j$, their partial order relation $\leqslant$ can be described as:

$$(A_i, B_i) \leqslant (A_j, B_j) \Leftrightarrow A_i \subseteq A_j \Leftrightarrow B_j \subseteq B_i \qquad (3)$$

A collection of all complete cognitive concepts of the formal context $L = (G, M, I)$ can be expressed as $C(K)$, while $(C(K), \leqslant)$ is a complete lattice, which is called a concept lattice [10].

**Definition 2** Formal decision context is a five tuple $K = (G, M, I, D, J)$, where $(G, M, I)$ and $(G, D, J)$ are formal contexts, $M$ is called conditional attribute set, $D$ is called decision attribute set. If $L(G, M, I) \leqslant L(G, D, J)$, then $K$ is called consistent, otherwise it is called inconsistent.

*2.2. Three-way decision*

The core idea of 3WD theory is to divide a general field into three pairs of disjoint sub fields. It provides a unique way of thinking for human problem solving and information processing [9, 30, 37]. Because of its simplicity and practicability, the principle and method of 3WD are widely used in all aspects of life [5, 22, 23, 36, 38, 39]. Yao gave the formal definition of three-way decision as follows [27, 28]:

**Definition 3** Let $U$ be a finite nonempty set and $(P, \leq)$ a totally ordered set. For any $\alpha$, $\beta \in P$, $\beta < \alpha$. Suppose that the set of designated values for acceptance is given by $P^+ = \{t \in P \mid t \geq \alpha\}$ and the set of designated values for rejection is given by $P^- = \{b \in P \mid b \leq \beta\}$. For an evaluation function $v : U \to P$, its three regions are defined by:

$$POS_{(\alpha, \beta)}(v) = \{x \in U \mid v(x) \geq \alpha\} \qquad (4)$$

$$NEG_{(\alpha, \beta)}(v) = \{x \in U \mid v(x) \leq \beta\} \qquad (5)$$

$$BN_{(\alpha, \beta)}(v) = \{x \in U \mid \beta < v(x) < \alpha\} \qquad (6)$$

For the binary information table called formal context in FCA, Qi et al. propose negative operators, three-way operators and three-way concepts that can reveal more information than formal concepts, and construct three-way concept lattices. Professor Yan integrates the idea of 3WD and granular computing into the theory of APOS, and explores the construction method of multi granularity three-way attribute partial order structure, which not only improves the efficiency of knowledge processing, but also makes the result of knowledge processing more reasonable [26].

For a formal context $K = (G, M, I, D, J)$, in the generation of causal attribute partial order structure, the strength of attribute causality is the standard for dividing objects $G$ and attributes $M$.

However, the decision attributes $D$ in the objects $G$ of the same layer is often not a pure positive domain (all 1) or a pure negative domain (all 0). More often, it is a boundary domain with both positive and negative attributes. At this time, for these boundary domains, it is necessary to delay judgment and improve the purity of the boundary domain through more attributes, which is the idea of 3WD.

### 2.3. Attribute partial order structure

In the theory of POFSA, given formal context $K = (G, M, I)$, for any attribute $m \in M$, the feature of attribute $m$ can be obtained by cognitive operator $(f : 2^M \to 2^G)$.

$$f(m) = \{g \mid gIm \ \forall g \in G\} \tag{7}$$

Based on the characteristics of each attribute of formal context, various attributes can be defined, such as maximum common attribute, mutually exclusive attribute, derivative attribute, super attribute, neighborhood derivative attribute and neighborhood super attribute.

There is a binary partial order relationship between derivative attributes and super attributes. According to the attribute partial order set $(M, \preccurlyeq)$, a hierarchical structure can be formed, which is called APOS. The Hasse diagram of this structure is called APOSD.

Based on the characteristics of each object in the formal context, various objects can also be defined, such as maximum common object, unique object, mutually exclusive object, etc. Similarly, based on the binary partial order relationship between derivative objects and super objects, the object partial order set $(U, \preccurlyeq)$ can form another hierarchical structure, which is called the object partial order structure (OPOS). The structure of Hasse diagram is called object partial order structure diagram (OPOSD).

From the perspective of knowledge visualization, APOSD represents the hierarchical relationship between attributes and the association relationship between objects, and OPOSD represents the hierarchical relationship between objects and the association relationship between attributes. We are chiefly concerned with causality of attributes, so in the next section, it starts from the perspective of APOSD.

### 2.4. Causal inference

In the era of high development of artificial intelligence, artificial intelligence has played a vital role in medical treatment, production and other fields. It has replaced human beings as the main force in some special fields. However, the current deep learning is only a special function $f(x)$ that can fit the data, or a polynomial distribution. It focuses on the relationship between the data label and the features to be learned. This relationship is often unreliable and has confused pseudo correlation [8, 14]. Judea pearl proposes in *the book of why* that deep learning not only needs a super fitting function, but also needs to deduce the internal causality from the logical relationship of facts.

CI is a subject proposed to solve the confusion in machine learning [2, 6]. Through CI, the influence of confusion can be reduced, so as to explore the essence of things. The process of reasoning is not only the process of eliminating confusion, but also the process of discovering things, recognizing things and mining cause effect relationships.

Causality is a fundamental notion in science, and plays an important role in explanation, prediction, decision-making, and control [7]. Judea pearl divides causality into three levels. From the bottom to the top are association, intervention and counterfactuals. The lowest level is association, which is also what deep learning is doing at present. That is to find out the correlation between variables through the observed data. Association cannot infer the direction in which events affect each other [14, 20]. For example, when event A occurs, event B also occurs, $P(B|A)$. But we can't find out that A leads to B. We only know that the two are related. The second level is intervention. Do operator is a commonly used method in CI to represent intervention and evaluate causality [15]. Through intervention, we can know whether event B will change with event A, $P(B|do(A))$. The highest level is counterfactuals [16], which can also be understood as "cause of effect". If B changes, can it be realized by changing event A.

CI pays more attention to the causal relationship between data when processing data, and blocks the path from confusion to features by means of intervention, so as to improve the causality between data. In this paper, we use the second level of causality - intervention to discover knowledge.

## 3. Method

In this section, we propose a method to construct 3WCAPOS which is a structure constructed by the attribute causality in formal decision context. In 3WCAPOS, with the increase of the number of structural layers, the object set gradually changes from fuzzy domain to pure positive domain or pure negative domain, which makes the structure of knowledge clearer.

The overall process of this method is shown in Fig. 1. CF blocks the path of confusion to attribute $m$ in directed acyclic graph by means of intervention, and it reduces the impact of confusion. The CF is used to evaluate the causal correlation degree between attribute $m$ and decision attribute $c$ in the formal decision context. On this basis, combined with the idea of 3WD, 3WCAPOS is formed.

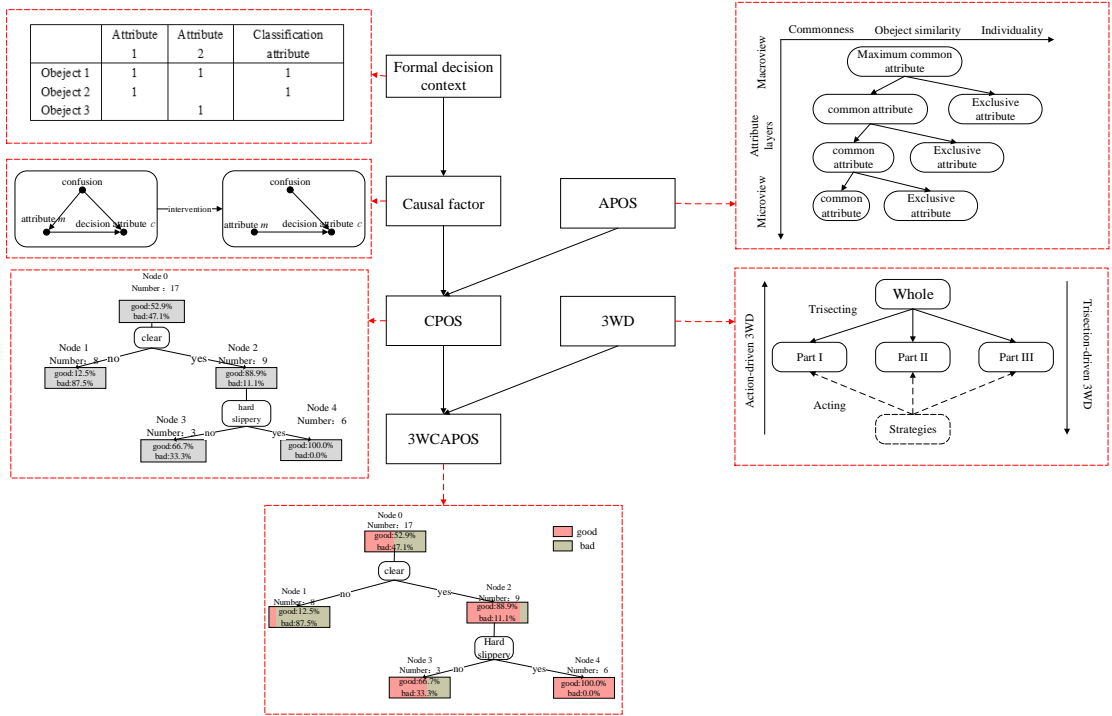

**Fig. 1.** The overall architecture of the method in this paper

### 3.1. Causal factor

In a formal decision context $K = (G, M, I, D, J)$, in order to describe the causal correlation degree of attribute $m_i \in M$ under decision attribute $c \in D$, we put forward the concept of casual factor(CF).

**Definition 4** In a formal decision context $K = (G, M, I, D, J)$, $\forall m_i \in M$, $\forall c \in D$, causal correlation degree is used to measure the causality of attribute $m_i$ to decision attribute $c$. The stronger the degree of causality, the greater the impact of representative attribute $m_i$ on decision attribute $c$. The weaker the causal correlation degree, the smaller the impact of representative attribute $m_i$ on decision attribute $c$.

**Definition 5** In a formal decision context $K = (G, M, I, D, J)$, $\forall m_i \in M$, $\forall c \in D$, if attribute $m_i$ has causal correlation with decision attribute $c$, then $m_i$ is the certain causal attribute under the condition of $c$; If attribute $m_i$ has no causal correlation with decision attribute $c$, then $m_i$ is an uncertain causal attribute under the condition of $c$.

**Example 1.** Table 1 shows the formal decision context transformed from watermelon dataset 2.0 in the field of machine learning. In this paper, black, curled, turbid, clear, concave and hard slippery are taken as the attributes of the dataset in Table 1. The value of the row object with this column attribute is 1, otherwise it is 0 (not filled here), so as to convert the watermelon dataset into the formal decision context.

**Table 1**

Formal decision context of watermelon dataset.

|     | black | curled | turbid | clear | concave | hard slippery | good |
| --- | --- | --- | --- | --- | --- | --- | --- |
| 1   |     | 1   | 1   | 1   | 1   | 1   | 1   |
| 2   | 1   | 1   |     | 1   | 1   | 1   | 1   |
| 3   | 1   | 1   | 1   | 1   | 1   | 1   | 1   |
| 4   |     | 1   |     | 1   | 1   | 1   | 1   |
| 5   |     | 1   | 1   | 1   | 1   | 1   | 1   |
| 6   |     |     | 1   | 1   |     |     | 1   |
| 7   | 1   |     | 1   |     |     |     | 1   |
| 8   | 1   |     | 1   | 1   |     | 1   | 1   |
| 9   | 1   |     |     |     |     | 1   |     |
| 10  |     |     |     | 1   |     |     | 1   |
| 11  |     |     |     |     |     | 1   |     |
| 12  |     | 1   | 1   |     |     |     |     |
| 13  |     |     | 1   |     | 1   | 1   |     |
| 14  |     |     |     |     | 1   | 1   |     |
| 15  | 1   |     | 1   | 1   |     |     |     |
| 16  |     | 1   | 1   |     |     | 1   |     |
| 17  |     | 1   |     |     |     | 1   |     |

**Definition 6** In a formal decision context $K = (G, M, I, D, J)$, $\forall m_i \in M$, $\forall c \in D$, if $p(c \mid m_i) = 0$, it is proved that $m_i$ and $c$ are not related, if there is no correlation, there is no causal correlation. If $p(c \mid m_i) \neq 0$, it is proved that $m_i$ and $c$ are related. Here the $CF$ is defined as:

$$CF(m_i \mid c) = \frac{p(c \mid do(m_i))}{p(c \mid m_i)} \tag{8}$$

**Property 1.** In a formal decision context $K = (G, M, I, D, J)$, $\forall m_i \in M$, $\forall c \in D$, CF has the following properties:

1. There is no causal correlation between $m_i$ and $c$, and $m_i$ is an uncertain causal attribute of $c$ when $CF = 1$.

2. $m_i$ is the ideal certain causal attribute of $c$ when $CF = 0$.

3. $m_i$ is the certain causal attribute of $c$, and the closer $CF$ is to 1, the weaker the causal correlation is, the closer $CF$ is to 0, the stronger the causal correlation is when $0 < CF < 1$.

4. $m_i$ is the certain causal attribute of $c$, and the closer $CF$ is to 1, the weaker the causal correlation is, the greater the $CF$, the stronger the causal correlation when $CF > 1$.

5. The greater the value of $\left| \log CF \right|$ corresponding to $m_i$, the stronger the causal correlation when $m_i$ is the certain causal attribute of $c$.

**Proof.**

1. If $CF = 1$, $CF(m_i | c) = p(c | do(m_i)) / p(c | m_i) = 1$, because $p(c | m_i) \neq 0$ and $0 < p(c | m_i) \leq 1$, $0 < p(c | do(m_i)) \leq 1$, so $p(c | do(m_i)) = p(c | m_i)$. There is no change between the two before and after the intervention. There is no causal correlation between $m_i$ and $c$. $m_i$ is the uncertain causal attribute of $c$.

2. If $CF = 0$, $CF(m_i | c) = p(c | do(m_i)) / p(c | m_i) = 0$, because $p(c | m_i) \neq 0$ and $0 < p(c | m_i) \leq 1$, $0 < p(c | do(m_i)) \leq 1$, so $p(c | do(m_i)) = 0$. Intervention makes $m_i$ and $c$ become irrelevant from relevant, $m_i$ is the ideal causal attribute of $c$.

3. If $0 < CF < 1$, $0 < CF(m_i | c) = p(c | do(m_i)) / p(c | m_i) < 1$, because $p(c | m_i) \neq 0$ and $0 < p(c | m_i) \leq 1$, $0 < p(c | do(m_i)) \leq 1$, so $p(c | do(m_i)) < p(c | m_i)$. Intervention reduces the influence of attribute $m_i$ on $c$, which proves that there is a causal correlation between attribute $m_i$ and $c$, and $m_i$ is the certain causal attribute of $c$.

4. If $CF > 1$, $CF(m_i | c) = p(c | do(m_i)) / p(c | m_i) > 1$, because $p(c | m_i) \neq 0$ and $0 < p(c | m_i) \leq 1$, $0 < p(c | do(m_i)) \leq 1$, so $p(c | do(m_i)) > p(c | m_i)$. Intervention increases the influence of attribute $m_i$ on $c$, which proves that there is a causal correlation between attribute $m_i$ and $c$, and $m_i$ is the certain causal attribute of $c$.

5. If $0 < CF < 1$, The smaller the value of $CF$, the greater the change caused by intervention,

that is, the higher the causal correlation between $m_i$ and $c$. Similarly, when $CF > 1$, the greater the value of $CF$, the greater the change caused by intervention, that is, the higher the causal correlation between $m_i$ and $c$. To sum up, when $m_i$ is the deterministic causal attribute of $c$, the greater the $|\log CF|$ value corresponding to $m_i$, the stronger the causal correlation.

For formal decision context of watermelon dataset in Table 1, the CF of attribute clear is evaluated as $CF(\text{clear} | \text{good}) = p(\text{good} | do(\text{clear})) / p(\text{good} | \text{clear}) = (1/8)/(8/9) \approx 0.141$.

Other properties are evaluated in the same way. After calculation, the causal correlation degree of all attributes to the decision attribute (good melon) are shown in the table below.

**Table 2**

Causal correlation ranking of watermelon dataset.

| attribute $m$ | $CF(m|\text{good})$ | $|\log(CF(m|\text{good}))|$ |
|---|---|---|
| clear | 0.141 | 1.962 |
| concave | 0.560 | 0.580 |
| black | 0.682 | 0.383 |
| curled | 0.711 | 0.341 |
| turbid | 0.714 | 0.336 |
| hard slippery | 1.200 | 0.182 |

Since the value range of $CF$ is $[0, +\infty)$ and the value range of causal correlation degree $|\log CF|$ is $[0, +\infty)$, in order to facilitate the study of the nature of causal correlation degree, we limit the value of causal correlation degree to a fixed interval by the following normalization means, and the normalized causal correlation degree is recorded as NC (Normalized Causality).

**Definition 7** Given a function $f(x)$, normalize the $CF$. The value range of the processed $CF$ is $(y_{\min}, y_{\max})$, $y_{\max}$ represents the upper bound of the $CF$ and $y_{\min}$ represents the lower bound of the $CF$.

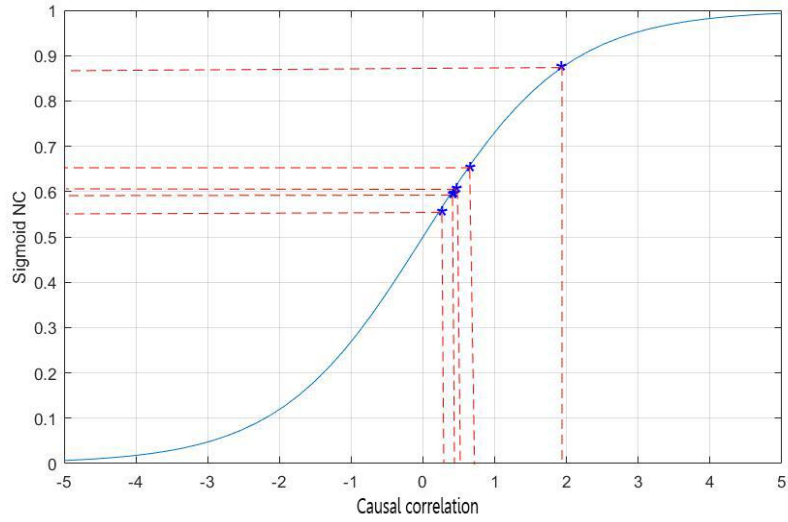

**Fig. 2.** Sigmoid normalized NC value range diagram (* indicates the value corresponding to watermelon dataset attribute. The value range of causal correlation degree is $[0,+\infty)$, after sigmoid normalization, its value range becomes $[1/2,1)$, which makes an infinite interval into a finite interval.)

When $f(x)$ takes sigmoid function, $f(x)=1/(1+e^{-x})$, the normalized *CF* value range is shown in Fig. 2. At this time, the upper bound $y_{max}$ of *CF* is 1 and the lower bound $y_{min}$ of CF is 0.

After normalization of sigmoid function, the value range of *NC* is $[1/2,1)$, where $1/2$ is the lower bound, and the lower bound is the point without causal correlation degree. And 1 is the upper bound, upper bound is the point with the strongest causal correlation. The closer to the lower bound, the weaker the causal correlation of the attribute. The closer to the upper bound, the stronger the causal correlation of the attribute.

**Table 3**

Sigmoid *NC* in watermelon dataset.

| Attribute $m$ | $|\log(CF(m|good))|$ | Sigmoid *NC* |
|---|---|---|
| clear | 1.962 | 0.877 |
| concave | 0.580 | 0.641 |
| black | 0.383 | 0.595 |
| curled | 0.341 | 0.584 |
| turbid | 0.336 | 0.583 |
| hard slippery | 0.182 | 0.545 |

*NC* of watermelon dataset after sigmoid normalization is shown in Table 3, the values of all attributes are in $[1/2,1)$, and the clear attribute is the attribute with the strongest causal correlation degree, which should be located at the top of the APOS. Hard slippery attribute is the attribute with the weakest causal correlation and should be located at the lowest level of APOS.

*3.2. Causal partial order structure*

**Definition 8** In a formal decision context $K = (G, M, I, D, J)$, $\forall m_i \in M$, $\forall c \in D$, by calculating the normalized causal correlation degree $NC(m_i \mid c)$, the causal relationship between attribute $m_i$ and decision attribute $c$ is determined according to the size of $NC(m_i \mid c)$, and the resulted partial order structure is causal partial order structure (CPOS).

In the formal context of CPOS, in addition to the concept of formal decision context, which is the data basis for attribute exploration, the causal coverage of attributes is also a necessary definition for constructing causal partial order structure diagram. In the formal decision context $K = (G, M, I, D, J)$, $\forall m_i \in M$, $\forall c \in D$, the causal coverage of attributes can be obtained by $NC$ through function $f(x)$.

$$NC(m_i \mid c) = f\left( \left| \log\left( \frac{p(c \mid do(m_i))}{p(c \mid m_i)} \right) \right| \right) \tag{9}$$

**Definition 9** In a formal decision context $K = (G, M, I, D, J)$, if $m_i, m_j \in M$ and $c \in D$, the $NC$ of $m_i$ is greater than that of $m_j$, then $m_i$ is believed said to have causal coverage over $m_j$.

$$NC(m_i \mid c) > NC(m_j \mid c) \tag{10}$$

**Property 2.** If A has a causal covering relationship with B, A is located in the upper layer of B in the causal partial order diagram.

**Proof.** In this paper, *CF* is used as the generation rule of the partial order graph. Attribute with greater causal coverage appears in the upper layer of the partial order graph.

**Example 2.** If $NC(m_i \mid c)$ of attribute $m_i$ is equal to 0.8 and $NC(m_j \mid c)$ of attribute $m_j$ is equal to 0.3, the relative relationship between them is shown in Fig. 3.

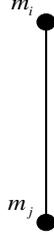

**Fig. 3.** Relative position diagram of $m_i$ and $m_j$ in CPOS

**Definition 10** In a formal decision context $K = (G, M, I, D, J)$, for attribute $m_i \in M$, if $CF(m_i)$ satisfies formula (11), then $m_i$ is called maximum causal coverage attribute or focus causal attribute.

$$CF(m_i \mid c) = 0 \tag{11}$$

**Property 3.** If attribute $m_i$ is the maximum causal coverage attribute or focus causal attribute, in the causal partial order structure, $m_i$ is at the top of the current remaining attributes.

**Proof.** According to item 2 in Property 1, when the $CF$ value of the current attribute $m_i$ is 0, the attribute is the ideal certain causal attribute. The ideal certain causal attribute is the attribute with the strongest causal correlation in the attribute set, so it is located at the top of the remaining attributes.

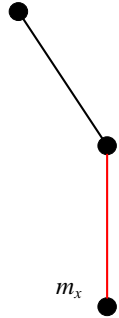

**Fig. 4.** Conditional causal partial order structure diagram of $m_x$ (Both ends of the black line are the existing conditional node $m_1$, $m_2$, and the bottom $m_x$ of the red line is a new node.)

**Definition 11** In a formal decision context $K = (G, M, I, D, J)$, if $m_1, m_2, m_i \in M$ and $c \in D$, under the condition of attributes $m_1$ and $m_2$, the conditional $CF$ of $m_x$ is $CF(m_x \mid m_1, m_2, c)$, and the normalized conditional causal correlation degree is $NC(m_x \mid m_1, m_2, c)$.

**Example 3.** If the current layer of the causal partial order structure graph is under the conditions of attributes $m_1$ and $m_2$, and $m_x$ is the maximum causal coverage attribute or focus causal attribute, the conditional causal partial order relationship of $m_x$ is shown in Fig. 4.

**Definition 12** In a formal decision context $K = (G, M, I, D, J)$, $\forall c \in D$, according to the purity of decision attribute $c$ in object set $G$, it can be divided into pure positive domain, pure negative domain and boundary domain. The object set whose values of decision attribute $c$ are all 1 is pure domain, the object set whose values of decision attribute $c$ are all 0 is pure negative domain, and the object set whose values of decision attribute $c$ are both 1 and 0 is boundary domain.

Based on APOS, we introduce the concept of CF to form CPOS. APOS determines the position of attributes in CPOS by calculating the *CF* value of attributes, and transforms the partial order structure from set coverage to causal coverage, which improves the interpretability and causality of the structure.

### 3.3. Three-way causal attribute partial order structure

On the basis of CPOS, this section integrates the idea of 3WD to form 3WCAPOS. In a formal decision context $K = (G, M, I, D, J)$, where $m_i \in M$ and $c \in D$, by calculating the normalized causal correlation degree $NC(m_i \mid c)$, the causal relationship between attribute $m_i$ and decision attribute $c$ is determined according to the size of $NC(m_i \mid c)$. On this basis, the purity of each node is evaluated to generate 3WCAPOS. Compared with CPOS, the purity of each node in the partial order structure is more obvious due to the addition of 3WD. At the same time, it also shows how the purity of 3WCAPOS generated by CF improves with the increase of structure level, and it finally reaches the pure positive domain, pure negative domain or boundary domain that cannot be divided by object set.

### 3.3.1. Data preprocessing method

Since 3WCAPOS can only process binary attributes and binary classification datasets, it is necessary to process datasets that are not binary attributes into binary attributes before constructing 3WCAPOS. The processing methods are as follows:

For discrete multivalued attributes, it is easier to convert them to binary attributes. Given a discrete multivalued attribute $a$, the set of values is $V_a = \{v_1, v_2, ..., v_n\}$. A new set of binary attributes $A^T$ can be obtained by $A^T = a \times V_a = \{a\_v_1, a\_v_2, ..., a\_v_n\}$.

For continuous attributes, conversion to binary attributes can be obtained by Algorithm 1.

---

**Algorithm 1：** Convert continuous attributes to binary attributes

---

**Input：**

  Continuous attribute $b$ and its value set $V_b = \{v_1, v_2, ..., v_n\}$

**Output：**

  Continuous attribute b bisection interval $B^T = \{[o_1, s_m], [s_m, o_n]\}$

1: Sort $V_b = \{v_1, v_2, ..., v_n\}$ and get ordered set $O_b = \{o_1, o_2, ..., o_n\}$ ($o_1$ is the minimum value, $o_n$

  is the maximum value)

2: **for** each $o_i$ in $O_b$

3:  $s_i = \left(\left(o_i + o_{i+1}\right)/2 \mid 1 \le i \le n-1\right)$

4: **end**

5: Get a set $S_b = \{s_1, s_2, ..., s_{n-1}\}$

6: **for** each $s_i$ in $S_b$

7:  **if** $v_i$ in $[o_1, s_i)$

8:   **then** make $v_i = 0$

9:  **end**

10: **if** $v_i$ in $[s_i, o_n]$

11:   **then** make $v_i = 1$

12: **end**

13:  Compute causal correlation $\left(s_i, |\log CF|_i\right)$

14: **end**

15: Get a normalized causal correlation set $R = \{(s_1, NC_1), (s_2, NC_2), ..., (s_{n-1}, NC_{n-1})\}$

16: Get max normalized causal correlation $NC_m$ and its $s_m$

17: **Return** a new bisection interval $B^T = \{[o_1, s_m), [s_m, o_n]\}$

---

*3.3.2. Construction method of three-way causal attribute partial order structure*

  In a formal decision context $K = (G, M, I, D, J)$, where $m_i \in M$ and $c \in D$, based on the method in Section 3.3.1, it is transformed into a formal context $K' = (G, M, I, D, J)$ that meets the generation requirements, and then 3WCAPOS is constructed through Algorithm 2 from the rules in Section 3.2:

---

**Algorithm 2**：Construction method of Three-way causal attribute partial order structure

---

**Input**：

Appropriate formal context $K' = (G, M, I, D, J)$

**Output**：

Three-way causal attribute partial order structure

1: **for** each $m_i$ in $M$

2:    Compute $CF(m_i | c) = (P(c | do(m_i)) / P(c | m_i) | m_i \in M, c \in D)$

3: **end**

4: Get a casual factory set $CFL = \{CF(m_i | c) | m_i \in M, \ c \in D\}$

5: Compute the normalized set $CFL$ and sort it as $CFL'$

6: Select the maximum in set $CFL'$ and its corresponding attribute $m_0$ as best partition

   attribute and $m_0$ is the node of structure

7: Divide the object set $G$ into two parts $G_n$ and $G_p$

8: Delete attribute $m_0$ in attribute set $M$ of $K' = (G, M, I, D, J)$

9: Repeat step 1-5 until no boundary object set exists or object set **G** cannot be partitioned

10: Return Three-way causal attribute partial order structure

---

The 3WCAPOS diagrams according to Example 1 are as follows:

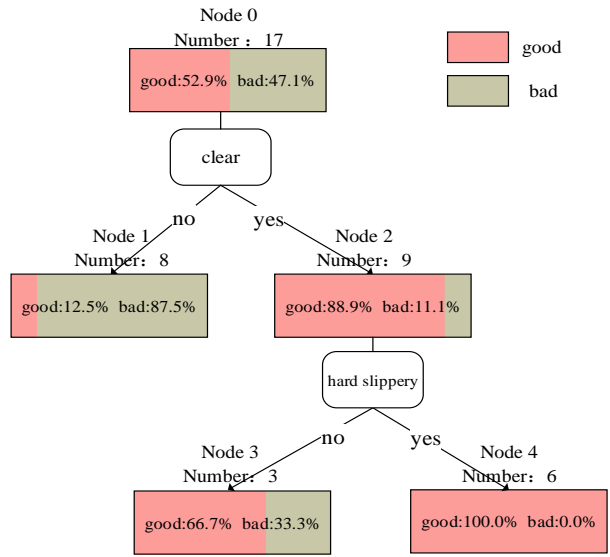

**Fig. 5.** 3WCAPOS diagram of watermelon dataset

According to the above algorithm, 3WCAPOS diagrams are generated based on the formal decision context of Example 1, which is shown in Fig. 5. In level 0, attribute clear is the attribute with the most extensive causal coverage and the most causal attribute in the original dataset. On the premise of clear attributes, node 1 and node 2 of level 1 are obtained according to whether the object contains clear attribute. The bad melon proportion of node 1 is 87.5%. The good melon proportion of node 2 is 77.8%. After the calculation of the remaining data, the classification can be continued, and the best attribute is hard slippery. In Level 2, node 3 and node 4 are obtained from the presence or absence of the hard slippery attribute in node 2. The sample number of node 3 is too small to be evaluated, and the purity of node 4 has reached 100%. So far, the construction of 3WCAPOS based on watermelon dataset has been completed.

## 4. Experiment

In this section, the construction method of 3WCAPOS proposed in Section 3.3 will be evaluated through six datasets from the UCI machine learning knowledge base. According to the different objects and methods of experimental comparison, this experiment is divided into two parts: precision comparison experiment and knowledge structure extraction comparison experiment. The precision comparison experiment is compared with the classification accuracy of CART, and the knowledge structure extraction comparison experiment is compared with the APOS. The arrangement of this section is as follows: Section 4.1 introduces the dataset and the results of data preprocessing. Section 4.2 is the precision comparison experiment. Section 4.3 is the comparative experiment of knowledge structure extraction.

### 4.1. Experimental data

Since 3WCAPOS only processes binary classification datasets, we select six UCI binary classification datasets. As shown in Table 5, there are simple Balloons Datasets with binary attributes (https://archive.ics.uci.edu/ml/datasets/Balloons). There are Early Stage Diabetes Risk Prediction Dataset attributes (https://archive.ics.uci.edu/ml/datasets/ Early+stage+diabetes+risk+ prediction+dataset) and Caesarean Section Classification Dataset (https://archive.ics.uci.edu/ml/ datasets/Caesarian+Section + classification + dataset) with both continuous and discrete attributes. There are Haberman's Survival Data Sets (https://archive.ics.uci.edu/ml/datasets/Haberman%27s +Survival), Wine Data Set (https://archive.ics.uci.edu/ml/datasets/Wine) and Parkinson's Dataset

(https://archive.ics.uci.edu/ml/datasets/Parkinsons) with only continuous attributes. The datasets are from simple to complex, which can fully prove the applicability and feasibility of this method.

Being similar to the APOS, 3WCAPOS can only analyze formal context, while the original datasets except Balloons Datasets are multivalued. Therefore, the first thing to do is data preprocessing: converting multi valued datasets into formal contexts.

**Table 4**

Description of experimental datasets in UCI database.

| Number | Balloons | Haberman's Survival | Wine | Diabetes | Caesarean | Parkinson |
|---|---|---|---|---|---|---|
| samples | 20 | 306 | 178 | 521 | 80 | 196 |
| original discrete attributes | 4 | 0 | 0 | 15 | 4 | 0 |
| original continuous attributes | 0 | 3 | 13 | 1 | 1 | 22 |
| binary attributes after conversion | 4 | 3 | 13 | 16 | 12 | 22 |

The above six datasets are preprocessed according to the data preprocessing method in Section 3.3.1. The number of discrete attributes and continuous attributes before and after conversion of each dataset is shown in Table 4.

### 4.2. Precision comparison experiment

In order to verify the classification and prediction performance of 3WCAPOS, we compare it with the classical machine learning algorithm classification and regression tree (CART). The structure of CART is similar to that of 3WCAPOS proposed in this paper. They are both tree structures, which are only different in model generation, so they are comparable.

### 4.2.1. Evaluating metrics

In order to accurately evaluate the classification model, we use five metrics commonly used in machine learning: accuracy (ACC), recall (REC), false positive rate (FPR), precision (PRE) and F1 Score. The five metrics are introduced as follows:

*ACC* is the proportion of the total number of correct predictions. The *ACC* is evaluated by dividing the number of all correct predictions by the total number of datasets. *ACC* can intuitively measure the classification ability of the model. The higher the *ACC*, the better the classification ability of the model. The *ACC* can be obtained by the following formula:

$$ACC = \frac{TP + TN}{TP + FP + FN + TN}(\%)$$  (12)

*REC* is the proportion of positive samples classified as correctly predicted in all real positive samples. *REC* measures the recognition ability of the model to positive examples. The greater the *REC*, the better the performance of the model. *REC* can be obtained by the following formula:

$$REC = \frac{TP}{TP + FN}(\%)$$  (13)

*FPR* is the proportion of counterexamples incorrectly classified as positive examples. *FPR* measures the ability of the model to misclassify. The greater the *FPR*, the worse the performance of the model. *FPR* can be obtained by the following formula:

$$FPR = \frac{FP}{TN + FP}(\%)$$  (14)

*PRE* is the correct proportion of the predicted positive sample. The *PRE* is similar to the *ACC*, which measures the correct classification ability of the model for positive samples. The greater the *PRE*, the better the performance of the model. The *PRE* can be obtained by the following formula:

$$PRE = \frac{TP}{TP + FP}(\%)$$  (15)

*F*1 *Score* is a metrics used to measure the accuracy of binary classification model in statistics. *F*1 *score* is a comprehensive evaluation metric, which is the harmonic average of *PRE* and *REC*. The larger the *F*1 *Score*, the better the performance of the model. The *F*1 *Score* can be obtained by the following formula:

$$F1\ Score = \frac{2 \times REC \times PRE}{REC + PRE}(\%)$$  (16)

Through the above metrics, we can analyze the advantages and disadvantages of 3WCAPOS in predicting the formal decision context in all cases.

### 4.2.2. Precision comparison experimental results

In this section, 3WCAPOS and CART are analyzed by leaving one method for cross validation. According to the principle of leaving one method for cross validation, one sample should be taken as the test set and the other as the training set. Assuming that the number of samples in the dataset is n, n models are finally generated to verify the advantages and disadvantages of this method. Based on the above principles, the preprocessed binary formal decision context in Table 4 is generated into

3WCAPOS and CART respectively. The evaluation metrics of this section are as follows:

**Table 5**

Result analysis of 3WCAPOS and CART.

| Metrics | Method | Balloons | Diabetes | Wine | Haberman's Survival | Caesarean | Parkinson |
|---------|--------|----------|----------|------|---------------------|-----------|-----------|
| ACC | 3WCAPOS | 1.000 | 0.994 | **0.954** | **0.745** | 0.750 | **0.841** |
| | CART | 1.000 | 0.994 | 0.738 | 0.735 | **0.775** | 0.821 |
| REC | 3WCAPOS | 1.000 | 1.000 | **0.983** | 0.996 | 0.739 | 0.993 |
| | CART | 1.000 | 1.000 | 0.797 | **1.000** | **0.913** | 0.993 |
| FRP | 3WCAPOS | 0.000 | 0.015 | **0.070** | **0.150** | 0.235 | 0.425 |
| | CART | 0.000 | 0.015 | 0.429 | 0.250 | 0.412 | 0.508 |
| PRE | 3WCAPOS | 1.000 | 0.991 | **0.921** | **0.744** | **0.810** | **0.830** |
| | CART | 1.000 | 0.991 | 0.839 | 0.735 | 0.750 | 0.811 |
| F1 Sore | 3WCAPOS | 1.000 | 0.995 | **0.951** | **0.852** | 0.773 | **0.904** |
| | CART | 1.000 | 0.995 | 0.818 | 0.847 | **0.824** | 0.893 |

The results of the leave one method obtained according to the evaluation metrics in Section 4.3 is shown in Table 5. 3WCAPOS has the same metrics as the CART in the relatively simple balloon and diabetes datasets, and the $ACC$ is close to 100%. In the more complex Wine, Haberman's survival and Parkinson datasets, the performance of 3WCAPOS is slightly better than that of CART. In the complex caesarean section classification dataset, 3WCAPOS is slightly lower than the CART in $ACC$, $REC$ and $F1$ $Score$, but higher than CART in $FPR$ and $ACC$.

To sum up, in the comparison of the six datasets in this experiment, 3WCAPOS performs slightly better than CART. The fundamental reason is that 3WCAPOS uses the $CF$ as the main factor to construct the structure. The $CF$ mines the causal relationship between attribute $m_i \in M$ and decision attribute $c \in D$ in the formal decision context $K = (G, M, I, D, J)$, reducing the impact of confusing attributes on decision attribute. Thus, the causality and stability of the model are improved.

Because the generated structure is relatively large, this section intercepts part of 3WCAPOS (layer 4) and CART (layer 3) in caesarean section classification dataset. The generated graphics are shown in Fig. 6. Obviously, compared with the structure of CART, 3WCAPOS is more inclined to mine rare but representative samples in the data. For example, there are only three samples whose node 2 is older than 36.5 years old, but their decision attributes are all caesarean section. In node 4, there are only 2 samples with 4 number of caesarean sections, and their decision attribute are also caesarean section. In CART, these rare but important samples are ignored because they mine the data relevance, but 3WCAPOS composed of CF can capture the important causal information in the data and improve its importance level. The above differences can be applied to the analysis of rare cases in the medical field. The number of samples of rare diseases is generally rare compared with ordinary diseases, but some rare diseases may lead to serious health problems. Therefore, it is of practical significance to find rare but important samples in a large number of samples.

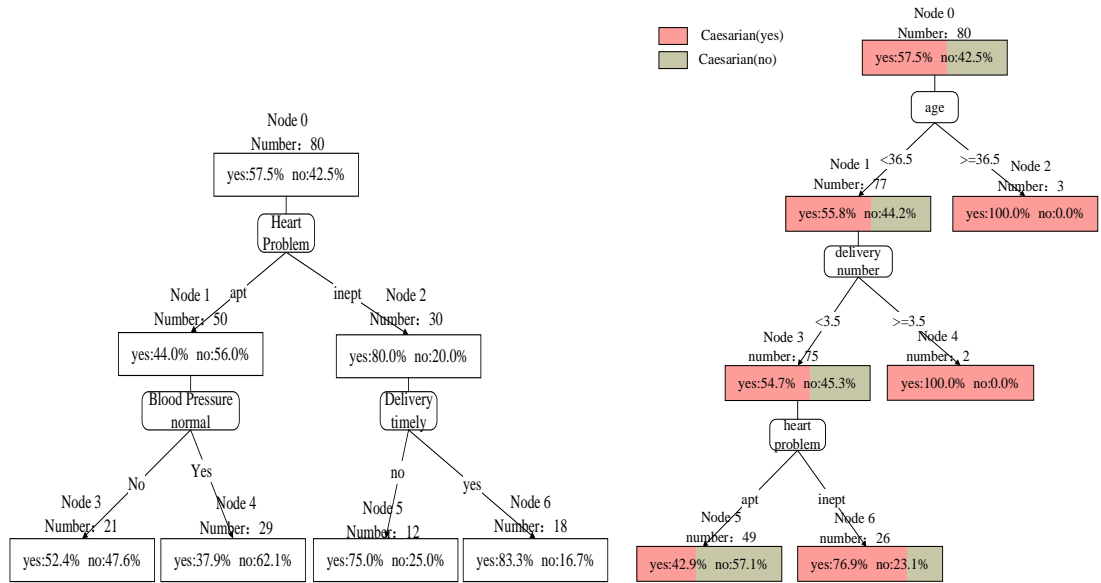

**Fig. 6.** Model structure comparison of caesarean section classification dataset

(Left: CART (part) Right: 3WCAPOS (part))

### 4.3. Comparative experiment of knowledge structure extraction

3WCAPOS is based on the improvement of APOS. By introducing the *CF*, the overall structure is evolved from the association relationship of frequent sets of objects to a more interpretable causal coverage relationship. In order to verify the structural advantages and disadvantages of 3WCAPOS, this section compares the balloon dataset, Haberman's survival dataset and caesarean section dataset. The three datasets are representative from simple to complex.

The balloon dataset has only four simple attributes: age, color, size, and action. It is the data used in cognitive psychology experiments. When the color is yellow and the size is small, it is inflated, otherwise it is not inflated. Through the balloon dataset, it can intuitively reflect whether the model can find the real reason for balloon inflation.

APOS and 3WCAPOS generated by the balloon dataset are shown in Fig. 7. The number of attributes in the APOS and the objects with the attribute are shown in Table 6. The number of objects in the dataset is 20, and each attribute accounts for 50% of the total number. Because no attribute m is satisfied $g(m) = U$, layer 0 is empty, and the set of attributes in layer 1 covers large, that is, frequent attributes, the four attributes are, the next layer is generated according to the APOS generation rules. The 3WCAPOS layer 0 is the attribute color. First, the decision attribute c is excluded as the non-inflated sample through the color attribute, and then inflated sample and the non-inflated sample can be found through the defecation of the attribute size. It can be seen that the leaf node purity of 3WCAPOS is 100%. From the structural analysis, the structure of 3WCAPOS is simpler, and the purity of decision attributes of samples is clear at a glance, while the structure of APOS is more complex and difficult to understand. From the analysis of generation rules, it can be seen from the description of the balloon dataset that the samples with yellow color and small size are inflated, and the rest are not inflated. 3WCAPOS uses CF to grasp the causality in the dataset, and only two important attributes are used to represent the relationship of the dataset, In the APOS, because it is based on frequent patterns and association relations, only attributes with a large number of samples are found. Such a structure is not easy to mine important information in the dataset.

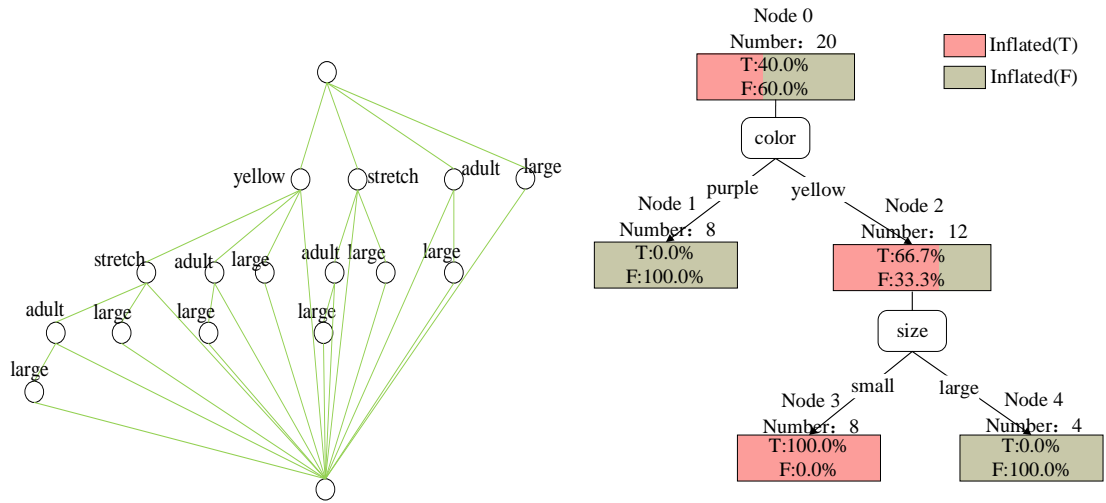

**Fig. 7.** Model structure comparison of balloon dataset
(Left: APOS Right: 3WCAPOS)

**Table 6**

Balloon dataset property and the number of objects that own it.

| Attribute $m$ | Number of objects $g(m)$ |
|---|---|
| yellow | 12 |
| large | 8 |
| stretch | 10 |
| adult | 10 |

The Haberman's Survival dataset has three consecutive attributes: age (patient's age at the time of surgery), year (patient's year of surgery), and number (number of positive axillary lymph nodes detected). From the description of the dataset, it can be seen that the long and short survival of patients has a stronger causal correlation with age and number, which can be used to test whether the model can mine this important information.

The APOS and 3WCAPOS generated by the Haberman's Survival dataset are shown in Fig. 8. The attributes in APOS and the number of objects with the attribute are shown in Table 7. The total number of objects in the dataset is 306. Because no attribute $m$ is satisfied $g(m) = U$, layer 0 is empty, layer 1 is all attributes, and the attribute year (105) with the largest number of objects continues to be generated according to the rules. In 3WCAPOS, node 0 excludes objects older than 77.5 (survival time less than 5 years) according to age attribute. Node 1 is the number (the number of positive axillary lymph nodes detected), and the last is node 4 - year (the year of operation). From the structural analysis, the APOS is relatively simple. There are only five branches to represent the dataset. Although the structure of 3WCAPOS is relatively complex, the structure is relatively clear. It can be seen that the purity of nodes is increased every time after the division of attribute nodes, from the initial mixed domain (node 0) to the pure negative domain (node 2 and node 5) and the pure positive domain (node 6). Although node 3 is a mixed domain, its purity has been increased compared with node 1. From the analysis of generation rules, 3WCAPOS uses CF to mine the causality in dataset. The order of causal correlation of attributes is age, number and year. The APOS is divided by the attribute year with the largest number of objects compared with other attributes in layer 1. According to the analysis of the dataset, the survival state of the patient has a stronger causal correlation with the patient's age and the number of positive axillary lymph nodes detected by the patient, but a weaker causal correlation with the year of operation. Therefore, from this perspective, the interpretability and mining ability of 3WCAPOS are better than APOS.

**Table 7**

The number of attributes in the Haberman's Survival dataset (after preprocessing) and the objects owning the attribute.

| Attribute $m$ | Number of objects $g(m)$ |
| --- | --- |
| Age (>=77.5) | 2 |
| Year (>=64.5) | 105 |
| Number (>=32.5) | 3 |

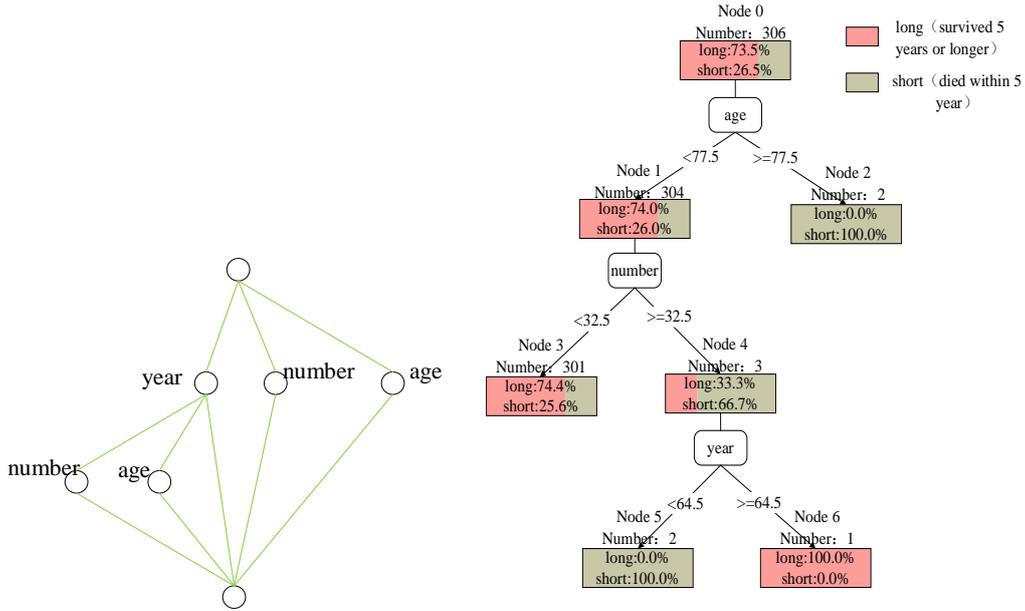

**Fig. 8.** Model structure comparison of Haberman's Survival dataset

(Left: APOS Right: 3WCAPOS)

Caesarean section classification dataset has both continuous and discrete attributes. It can be found from the observation dataset that the classification results of samples older than 36.5 years and more than 3.5 times of production are Caesarian, but the proportion of the number of samples of these two attributes in the total number of samples is very small. Therefore, the dataset can be used to test whether the model can mine rare but important information.

The APOS and 3WCAPOS (part) generated by the Caesarean section classification dataset are shown in Fig. 9. The attributes in the APOS and the number of objects with this attribute are shown in Table 8. The total number of objects in the dataset is 80. Because no attribute $m$ is satisfied $g(m) = U$, layer 0 is empty, and layer 1 is the attribute with more objects such as heart problem (50), delivery timely (46), normal blood pressure (40), low blood pressure (20) and high blood pressure (20) attributes. Then, the next layers are generated from these five attributes. While in layer 1, 3WCAPOS first divides samples into older than 36.5 or not. Then in layer 2, 3WCAPOS divides the remaining samples into delivery number greater than 3.5 or not. Next, 3WCAPOS considers remaining attributes such as heart problem and blood pressure and so on. From the structural analysis, both structures are complex, because the processed caesarean section dataset has 10 attributes, which means that both the structures need more layers to achieve the optimal structure. From the analysis of generation rules, the APOS tends to find attributes with a large number of samples as the nodes of the upper layer, while 3WCAPOS tends to find important attributes with high causal correlation. For example, elderly pregnant women or pregnant women with more than 3.5 births are rare but important samples. Therefore, 3WCAPOS puts them in the first or second layer, followed by heart attributes. From this point of view, the ability of 3WCAPOS to mine rare samples with strong causal correlation is greater than that of APOS.

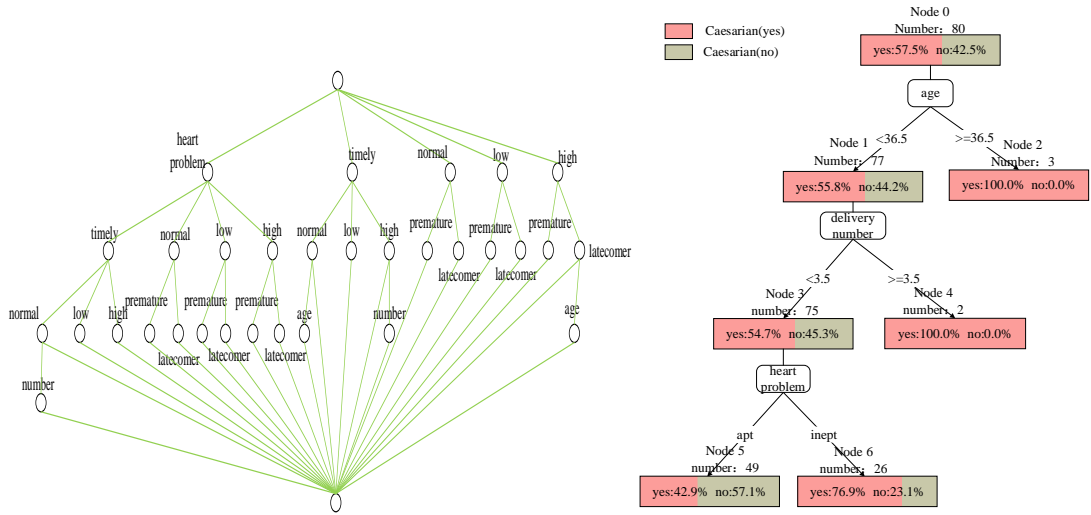

**Fig. 9.** Model structure comparison of Caesarean section classification dataset
(Left: APOS Right: 3WCAPOS (part))

**Table 8**

Attribute in Caesarean section classification dataset (after preprocessing) and the number of objects owning the attribute.

| Attribute $m$ | Number of objects $g(m)$ |
|---|---|
| Heart problem（inept） | 50 |
| Delivery timely | 46 |
| Delivery premature | 17 |
| Delivery latecomer | 17 |
| Blood pressure normal | 40 |
| Blood pressure low | 20 |
| Blood pressure high | 20 |
| Delivery Age（>=36.5） | 3 |
| Delivery Number（>=3.5） | 2 |

To sum up, through the comparison of three datasets, we can draw the following conclusions: the APOS is generated based on the frequent pattern of attributes. The attributes with more samples are often in the upper layer of the structure, and the attributes with less samples are in the lower layer of the structure, which shows the correlation relationship. While 3WCAPOS is generated based on CF, CF pays more attention to the causal correlation between attributes and decision attribute. Based on causal correlation, attributes and samples with strong causality in the data can be mined, especially for important attributes with few samples, and because of the upper layer of important attribute processing structure, the structure are clearer and more interpretable.

## 5. Conclusion

APOS is proposed under the situation of the rapid development of information technology. APOS can well solve the problems of cumbersome concept calculation and serious connection crossing in formal concept analysis and concept lattice, and can express the structural relationship between data attributes and the similar relationship between data objects. However, at this stage, the classical methods of constructing APOS are studied at the traditional set coverage level. The essence is frequent pattern and correlation analysis, which shows the data correlation of formal context. In this paper, we propose a metric for evaluating the causal correlation degree between attributes and decision attribute in the formal decision context, which is named CF. CF evaluates the causal correlation degree of attributes by means of intervention in causal reasoning, and limits it to a fixed interval after normalization for ranking the importance of attributes. Secondly, we define the concept of CPOS and its theorems, properties are proved to evaluate the causality between attributes and decision attribute. Then integrating the idea of three-way decisions makes the node information of CPOS clearer and trend of the structure more obvious. Finally, we propose a method of constructing 3WCAPOS based on CF, and prove the advantages of 3WCAPOS from two aspects based on six datasets: for one thing, compared with the CART with similar tree structure in accuracy, recall and F1 score, this structure is equal to or even slightly better than CART. For another, 3WCAPOS is structurally more interpretable and causal than CART and APOS, and is particularly sensitive to important attributes with few objects. In conclusion, through sufficient experimental comparison, it is proved that 3WCAPOS is superior to APOS in discovering causality between data and has a better classification ability.

The method proposed in this paper can be applied to the diagnosis and prediction of rare diseases, because rare diseases often have a small number of samples and are confused with some common diseases. This method can effectively mine the causal relationship in the data and mine the important features with a small number of samples. However, this method still has some shortcomings, such as the classification ability and troublesome processing of complex datasets still needs to be improved. The next research work is to combine CF with random forest in ensemble learning, hoping to further improve classification ability, structural stability and interpretability.

**CRediT authorship contribution statement**

**Zaifa Xue:** Investigation, Methodology, Writing-original draft, Validation. **Huibin Lu:** Supervision, Methodology, Funding acquisition. **Tao Zhang:** Conceptualization, Methodology, Project administration, Funding acquisition. **Tao Li:** Software, Validation, Writing. **Xin Lu:** Data curation, Writing-review & editing.

**Declaration of competing interest**

The authors declare that they have no known competing financial interests or personal relationships that could have appeared to influence the work reported in this paper.

**Acknowledgments**

This work is financially supported by the National Natural Science Foundation of China (under Grants 62176229), the Natural Science Foundation of Hebei Province (under Grants F2020203010), and the Humanities and Social Sciences Foundation of the Ministry of Education of China (under Grant no. 19YJA740076). The authors gratefully acknowledge the supports.